# Introduction to Rare-Event Predictive Modeling for Inferential Statisticians

– A Hands-On Application in the Prediction of Breakthrough Patents –

Daniel S. Hain[*] and Roman Jurowetzki

*Aalborg University, Department of Business and Management, IKE / DRUID, Denmark*

**Abstract:** Recent years have seen a substantial development of quantitative methods, mostly led by the computer science community with the goal to develop better machine learning application, mainly focused on predictive modeling. However, economic, management, and technology forecasting research has up to now been hesitant to apply predictive modeling techniques and workflows. In this paper, we introduce to a machine learning (ML) approach to quantitative analysis geared towards optimizing the predictive performance, contrasting it with standard practices inferential statistics which focus on producing good parameter estimates. We discuss the potential synergies between the two fields against the backdrop of this at first glance, "target-incompatibility". We discuss fundamental concepts in predictive modeling, such as out-of-sample model validation, variable and model selection, generalization and hyperparameter tuning procedures. Providing a hands-on predictive modelling for an quantitative social science audience, while aiming at demystifying computer science jargon. We use the example of "high-quality" patent identification guiding the reader through various model classes and procedures for data pre-processing, modelling and validation. We start of with more familiar easy to interpret model classes (Logit and Elastic Nets), continues with less familiar non-parametric approaches (Classification Trees and Random Forest) and finally presents artificial neural network architectures, first a simple feed-forward and then a deep autoencoder geared towards anomaly detection. Instead of limiting ourselves to the introduction of standard ML techniques, we also present state-of-the-art yet approachable techniques from artificial neural networks and deep learning to predict rare phenomena of interest.

---

[*]Corresponding author, Aalborg University, Denmark
Contact: dsh@business.aau.dk

# 1 Introduction

Recent years have seen a substantial development of machine learning (ML) methods, mostly led by the computer science community with the goal to increase the predictive power of statistical models. This progress ranges over a broad host of applications – from computer vision, speech recognition, synthesis, machine translation, only to name a few. Many models at the core of such ML applications resemble those used by statisticians in social science context, yet a paradigmatic difference persists. Social science researchers traditionally apply methods and workflows from inferential statistics, focusing on the identification and isolation of causal mechanisms from sample data which can be extrapolated to some larger population. In contrast, ML models are typically geared towards fitting algorithms that map some input data to an outcome of interest in order to perform predictions for cases there the outcome is not (yet) observed. While this has led to many practical applications in the natural sciences and business alike, the social sciences have until recently been hesitant to include ML in their portfolio of research methods. Yet, the recent availability of data with improvements in terms of quantity, quality and granularity Einav and Levin (2014a,b), led to various calls in the business studies (McAfee et al., 2012) and related communities for exploring potentials of ML methods for advancing their own research agenda.

This paper provides a condensed and practical introduction to ML for social science researchers which are commonly received training in methods and workflows from inferential statistics.. We discuss the potential synergies and frictions between the two approaches, and elaborate on fundamental ML concepts in predictive modeling, such as generalization via out-of-sample model validation, variable and model selection, and hyperparameter tuning procedures. By doing so, we demystify computer science jargon common to ML, draw parallels and highlight commonalities as well as differences with popular methods and workflows in infrerential statistics.

We illustrate the introduced concept by providing a hands-on example of "high-quality" patent identification.[1] We guide the reader through various model classes and procedures for data pre-processing, modelling and validation. Atfer starting of with model classes familiar to applied interential statisticians (Logit and Elastic Nets), we continues with nonparametric approaches less commonly applied in the social sciences (simple Classification Trees, Random Forest and Gradient Boosted Trees), gradually shifting the emphasis towards hyperparameter tuning and model architecture engineering.vInstead of limiting ourselves to the introduction of standard ML techniques, we also present state-of-the-art yet approachable techniques from artificial neural networks (ANN) and deep learning to predict rare phenomena of interest. Finally, we provide guidance on how to apply these techniques for social science research and point towards promising avenues of future research which could be enabled by the use of new data sources and estimation techniques.

---

[1]Often, the challenge in adapting ML techniques for social science problems can be attributed to two issues: (1) Technical lock-ins and (2) Mental lock-ins against the backdrop of paradigmatic contrasts between research traditions. For instance, many ML techniques are initially demonstrated at a collection of – in the ML and Computer Science – well known *standard datasets* with specific properties. For an applied statistician particularly in social science, however, the classification of Netflix movie ratings or the reconstruction of handwritten digits form the MNIST data-set may appear remote or trivial. These two problems are adressed by contrasting ML techniques with inferential statistics approaches, while using the non-trivial example of patent quality prediction which should be easy to comprehend for scholars working in social science disciplines such as economics.

# 2 An Introduction to Machine Learning and Predictive Modeling

## 2.1 Predictive Modeling, Machine Learning, and Social Science Research

Until recently, the social sciences where not overly eager to embrace and apply the methodological toolbox and procedural routines developed within the ML discipline. An apparent reason is given by inter-disciplinary boundaries and intra-disciplinary methodological "comfort zones" (Aguinis et al., 2009) as well as by path-dependencies, reinforced through the way how researchers are socialized during doctoral training (George et al., 2016). However, there also seems to be inherent – if not epistemological – tension between the social science inferential statistics and ML to data analysis, and how both could benefit from each other's insights is not obvious on first glance.

We here argue the ML community has not only developed "tricks" an social science statisticians in applied fields such as econometrics, sociometrics, or psychometrics might find extremely useful (Varian, 2014), but also enable new research designs to tackle a host of questions currently hard to tackle with the traditional toolbox of statistical inference. We expect such methods to broadly diffuse within quantitative social science research, and suggest the upcoming liaison of inferential statistics and ML to shake up our current routines. Here, highly developed workflows and techniques for predictive modelling appear to be among the most obvious ones.

In figure 1 we depict two trade-offs that we find relevant to consider in a paradigmatic discussion of data science and econometric approaches. On the one hand, and as presented in figure 1b, there is a general trade-off between the learning capacity of model classes and their interpretability. First, richer model classes are able to fit a complex functional form to a prediction problem, which improves their performance over simple (eg. linear parametric) models if (and only if) the underlying real-world relationship to model is equally complex. It can be assumed that this is the case for many relevant outcomes of the interaction in complex social systems (eg. economic growth, technology development, success of a start-up), and consequently, that richer models would have an edge for prediction tasks in such systems. With larger amounts of data available, such complex functional forms can be fitted more accurately, resulting in a higher learning capability of richer model classes compared to simple models which tend to saturate early, as illustrated in figure 1a.

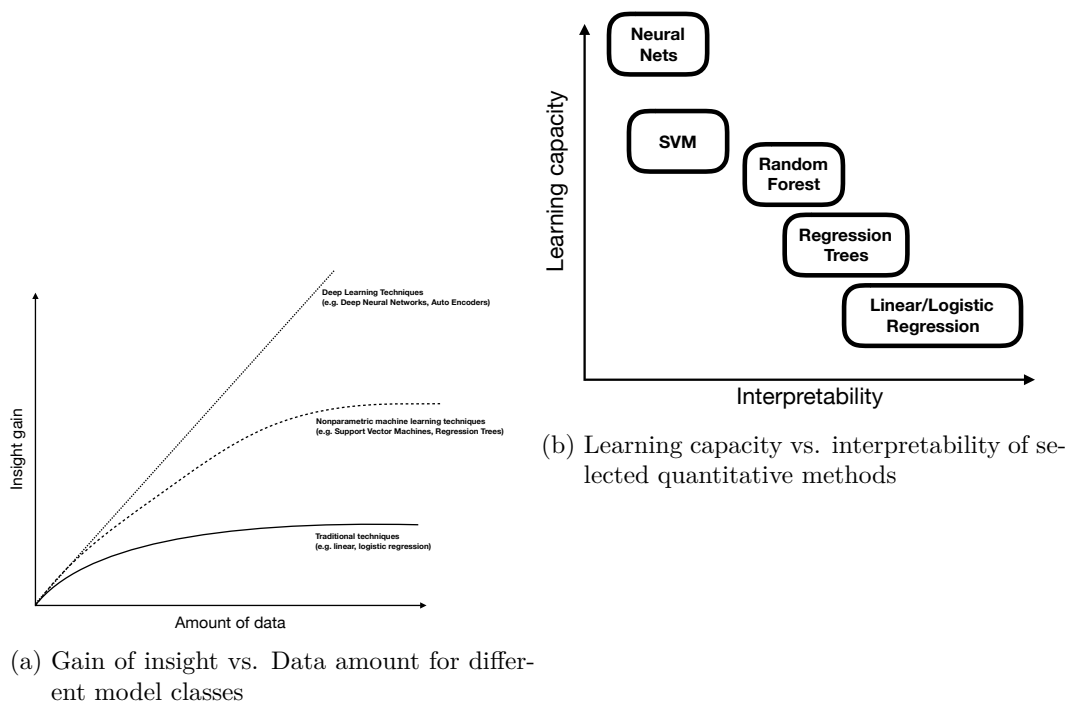

(a) Gain of insight vs. Data amount for different model classes

(b) Learning capacity vs. interpretability of selected quantitative methods

Figure 1: Learning capacity, amount of data and interpretability for different modeling techniques

The relationships between inputs and outputs captured by a linear regression model are easy to understand and interpret. As we move up and to the left in this chart, the learning capacity of the models increases. Considering the extreme case of deep neural networks, we find models that can capture interactions and nonlinear relations across large datasets, fitting in their complex functions between in- and outputs across the different layers with their multiple nodes. However, for the most part, it is fairly difficult if not impossible to understand the fitted functional relationship. This is not necessarily a problem for predictive modeling but of much use in cases where the aim is to find causal relationships between in- and outputs.

## 2.2 Contrasting Causal and Predictive Modeling

As applied econometricians, we are for the most part interested in producing good *parameter estimates*.[2] We construct models with unbiased estimates for some parameter $\beta$, capturing the relationship between a variable of interest $x$ and an outcome $y$. Such models are supposed to be "structural", where we not merely aim to reveal correlations between $x$ and $y$, but rather a causal effect of directionality $x \rightarrow y$, robust across a variety of observed as well as up to now unobserved settings. Therefore, we carefully draw from existing theories and empirical findings and apply logical reasoning to formulate hypotheses which articulate the expected direction of such causal effects. Typically, we do so by studying one or more bivariate relationships under *cetris paribus* conditions in a regression model, "hand-curate" with a set of supposedly causal variables of interest. The primary concern here is to minimize the standard errors $\epsilon$ of our $\beta$ estimates, the difference between our predicted $hat(y)$ and the observed $y$, conditional to a certain level of $x$, *ceteris paribus*. We are less interested in the overall predictive power of our model (Usually measured by the models $R^2$), as long as it is in a tolerable range.[3]

However, we are usually worried about the various type of endogeneity issues inherent to social data which could bias our estimates of $\beta$. For instance, when our independent variable $x$ can be suspected to have a bidirectional causal relationship with the outcome $y$, drawing a causal inference of our interpretation of $\beta$ is obviously limited. To produce unbiased parameter estimates of arguably causal effects, we are indeed willing to sacrifice a fair share of our models' explanatory power.

A ML approach to statistical modeling is, however, fundamentally different. To a large extent driven by the needs of the private sector, data analysis here concentrates on producing trustworthy predictions of outcomes. Familiar examples are the recommender systems employed by companies such as Amazon and Netflix, which predict with "surprising" accuracy the types of books or movies one might find interesting. Likewise, insurance companies or credit providers use such predictive models to calculate individual "risk scores", indicating the likelihood that a particular person has an accident, turns sick, or defaults on their credit. Instances of such applications are numerous, but what most of them have in common is that: (i.) they rely on a lot of data, in terms of the number of observations as well as possible predictors, and (ii.) they are not overly concerned with the properties of parameter estimates, but very rigorous in optimizing the overall prediction accuracy. The underlying socio-psychological forces which make their consumers enjoy a specific book are presumably only of minor

---

[2] We here blatantly draw from stereotypical workflows inherent to the econometrics and ML discipline. We apologize for offending whoever does not fit neatly in one of these categories.

[3] At the point where our $R^2$ exceeds a threshold somewhere around 0.1, we commonly stop worrying about it.

interest for Amazon, as long as their recommender system suggests them books they ultimately buy.

## 2.3 The Predictive Modeling Workflow

### 2.3.1 General idea

At its very core, in predictive modeling and for the most part the broader associated ML discipline, we seek for models and functions that do the best possible job in predicting some output variable $y$. This is done by considering some loss function $L(\hat{y}, y)$, such as the popular root-mean-square error (RMSE)[4] or the rate of missclassified observations, and then searching for a function $\hat{f}$ that minimizes our predicted loss $E_{y,x}[L(\hat{f}(x), y)]$.

To do so, the broader ML community has developed an enormeous set of techniques from traditional statistics but also computer science and other disciplines to tackle prediction problems of various nature. While some of those techniques are widely known and applied by econometricians and the broader research community engaged in causal modeling (e.g., linear and logistic regression) or lately started to recieve attention (e.g., elastic nets, regression- and classification-trees, kernel regressions, and to some extend random forests), others are widely unknown and rarely applied (e.g., support vector machines, artificial neural networks).[5] Figure A.1 attempts to provide a broad overview on popular ML model classes and techniques.

However, fundamental differences in general model building workflows and underlying philosophies makes the building as well as interpretation of (even familiar) predictive models with the "causal lense" of a trained econometrician prone to misunderstanding, misspecification, and misleading evaluation. Therefore, we in the following outlay some general principles of predictive modeling, before we in the following section illustrate them in an example.

First, in contrast to causal modeling, most predictive models have no *a priori* assumption regarding the direction of the effect, or any causal reason behind it. Therefore, predictive models exploit correlation rather that causation, and to predict rather than explain an outcome of interest. This provides quite some freedom in terms of which and how many variables (to introduce further ML jargon, henceforth called *features*) to select, if and how to transform them, and so forth. Since we are not interested in parameter estimates, we also do not have to worry so much about asymp-totic properties, assumptions, variance inflation, and all the other common problems in applied econometrics which could bias parameter estimates and associated standard errors. Since parameters are not of interest, there is also no urgent need to capture their effect properly, or have them at all. Indeed, many popular ML approaches are non-parametric and characterized by a flexible functional form to be fitted to whatever the data reveals. Equipped with such an arsenal, achieving a high explanatory power of a model appears quite easy, but every econometrician would doubt how well such a model generalizes. Therefore, without the limitations but also guarantees of causal modeling, predictive models are in need of other mechanisms to ensure their generalizability.

---

[4] As the name already suggest, this simply expresses by how much our prediction is on average off: $RMSE = \sqrt{\frac{\sum_{i=1}^{n}(\hat{y}_i - y_i)^2}{n}}$.

[5] Interestingly, quite some techniques associated with identification strategies which are popular among econometricians, such as the use of instrumental variables, endogenous selection models, fixed and random effect panel regressions, or vector autogregressions, are little known by the ML community.

### 2.3.2 Out-of-Sample validation

Again, as econometricians, we focus on parameter estimates, and we implicitly take their out-of-sample performance for granted. Once we set up a proper identification strategy that delivers unbiased estimates of a causal relationship between $x$ and $y$, depending on the characteristics of the sample, this effect supposedly can be generalized on a larger population. Such an exercise is *per se* less prone to over-specification since the introduction of further variables with low predictive power or correlation with $x$ tends to "water down" our effects of interest. Following a machine learning approach geared towards boosting the prediction of the model, the best way to test how a model predicts is to run it on data it was not fitted for. This can be done upfront dividing your data in a training sample, which you use to fit the model, and a test (or hold-out) sample, which we set aside and exclusively use to evaluate the models final prediction. This should only be done once, because a forth and back between tweaked training and repeated evaluation on the test sample otherwise has the effect of an indirect overfitting of the model.

Therefore, it is common in the training data also set a validation sample aside to first test the performance of different model configurations out-of-sample. Consequently, we aim at minimizing the *out-of-sample* instead of the *within sample* loss function. Since such a procedure is sensitive to potential outliers in the training or test sample, it is good practice to not validate your model on one single test-sample, but instead perform a *k-fold cross-validation*, where the loss function is computed as the average loss of $k$ (commonly 5 or 10)) separate test samples.[6] Finally, the best performing configuration is used to fit this model on the whole training sample. The final performance of this model is in a last step then evaluated by its prediction on the test sample, to which the model up to now has not been exposed to, neither direct nor indirect. This procedure is illustrated in figure 2.

---

[6]Such k-fold cross-validations can be conveniently done in R with the `caret`, and in Python with the `scikit-learn` package.

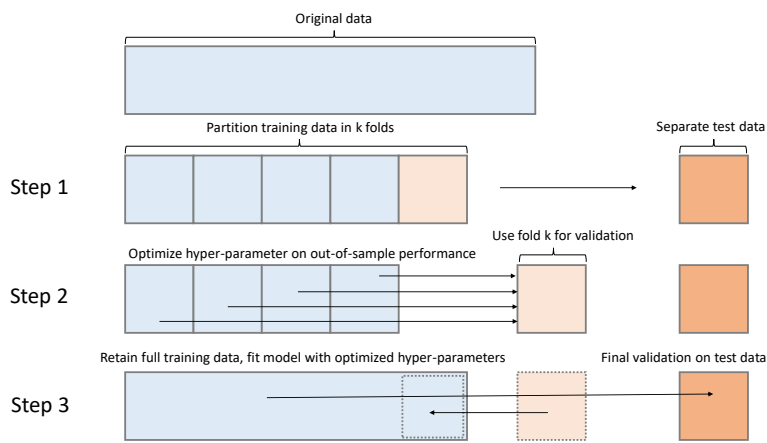

Figure 2: Intuition behind K-fold Crossvalidation

While out-of-sample performance is a standard model validation procedure in machine learning, it has yet not gained popularity among econometricians.[7] As a discipline originating from a comparably "small data" universe, it appears counterintuitive for most cases to "throw away" a big chunk of data. However, the size of data-sources available for mainstream economic analysis, such as register data, has increased to a level, where sample size cannot be taken anymore as an excuse for not considering such a goodness-of-fit test, which delivers much more realistic measures of a model's explanatory power. What econometricians like to do to minimize unobserved heterogeneity and thereby improve parameter estimates is to include a battery of categorical control variables (or in panel models, fixed effects) for individuals, sectors, countries, *et cetera*. It is needless to say that this indeed improves parameter estimates in the presence of omitted variables but typically leads to terrible out-of-sample prediction.

### 2.3.3 Regularization and hyperparameter tuning

Turning back to our problem of out-of-sample prediction, now that we have a good way of measuring it, the question remains how to optimize it. As a general rule, the higher the complexity of a model, the better it tends to perform within-sample, but also to loose predictive power when performing out-of-sample prediction. Since finding the right level of complexity is a crucial, researchers in machine learning have put a lot of effort in developing "regularization" techniques which penalize model complexity. In addition to various complexity restrictions, many ML techniques have additional

---

[7]However, one instantly recognizes the similarity to the nowadays common practice among econometricians to bootstrap standard errors by computing them over different subsets of data. The difference here is that we commonly use this procedure, (i.) to get more robust parameter estimates instead of evaluating the model's overall goodness-of-fit, and (ii.) we compute them on subsets of the same data the model as fitted on.

options, called *hyperparameter*, which influence their process and the resulting prediction. The search for optimal tuning parameters (in machine learning jargon called *regularization*, or hyperparameter tuning)[8] is at the heart of machine learning research efforts, somewhat its "secret sauce". The idea in it's most basic form can be described by the following equation, as expressed by (Mullainathan and Spiess, 2017):

Figure 3: In- vs. out-of-sample loss relationship

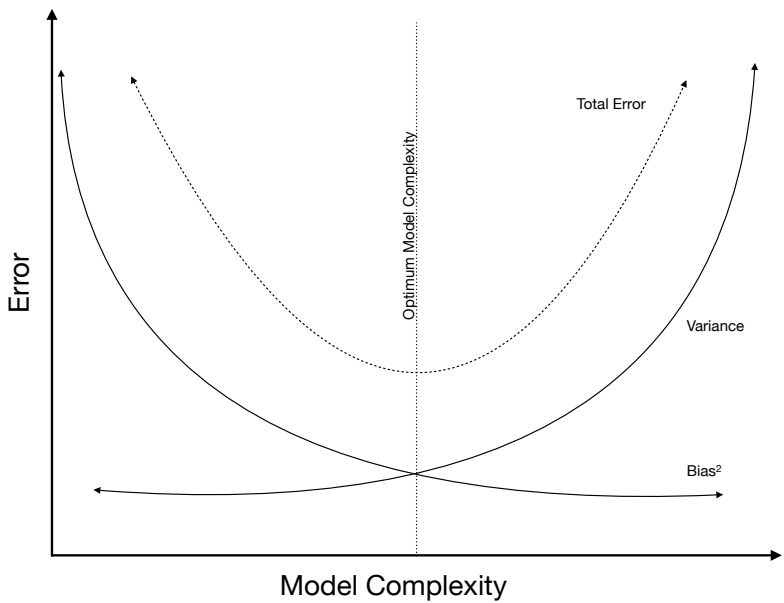

$$minimize \underbrace{\sum_{i=1}^{n} L(f(x_i), y_i)}_{in-sample\ loss},\ over\ \overbrace{f \in F}^{function\ class}\ subject\ to\ \underbrace{R(f) \leq c}_{complexity\ restriction}. \quad (1)$$

Basically, we here aim at minimizing the in-sample loss of a prediction algorithm of some functional class subject to some complexity restriction, with the final aim to minimize the expected out-of-sample loss. Depending on the technique applied, this can be done by either selecting the functions features $x_i$ (as we discussed before in "variable selection"), the functional form and class $f$, the complexity restrictions $c$, or other hyperparameters that influence the models internal processes. This process of model tuning in practice often is a mixture of internal estimation from the training data, expert intuition, and best practice, as well as trial-and-error. Depending on the complexity of the problem, this can be a quite tedious and lengthy process.

---

[8]For exhaustive surveys on regularization approaches in machine learning particularly focused on high-dimensional data, consider Pillonetto et al. (2014); Wainwright (2014).

The type of regularizations and model tuning techniques one might apply varies, depending on the properties of the sample, the functional form, and the the desired output. For parametric approaches such as OLS and logistic regressions, regularization is primarily centered around feature selection and parameter weighting. Many model tuning techniques are iterative, such as model *boosting*, an iterative technique involving the linear combination of prediction of residuals, where initially misclassified observations are given increased weight in the next iteration.

*Bootstrapping*, the repeated estimation of random subsamples, is in ML used primarily to adjust the parameter estimates by weighting them across subsamples (which is then called *bagging*).[9] Finally, *ensemble* techniques use the weighted combination of predictions done by independent models to determine the final classification.

## 3 An Application on Patent Data

In this section, we will illustrate the formerly discussed methods, techniques, and concepts at the example of PATSTAT patent data in order to develop a predictive model of high-impact (breakthrough) patents. In addition, we will "translate" necessary vocabulary differences between ML and econometrics jargon, and point to useful packages in R and Python, the current *de factor* standards for statistical programming in ML and also increasingly popular among applied econometricians. Here, our task is to predict a dichotomous outcome variable. In ML jargon, this is the simplest form of a *classification problem*, where the available classes are 0=no and 1=yes. As econometricians, probably our intuition would lead us to apply a linear probability (LPM) or some form of a logistic regression model. While such models are indeed very useful to deliver parameter estimates, if our goal is pure prediction, there exist much richer model classes, as we demonstrate in the following. The code as well as data used in the following exercise has been made publicly available under `https://github.com/ANONYMOUS` (altered for review).

### 3.1 Data and Context

#### 3.1.1 Context

Patent data has long been used as a widely accessible measure of inventive and innovative activity. Besides its use as an indicator of inventive activity, previous research shows that patents are a valid indicator for the output, value and utility of inventions (Trajtenberg et al., 1997), innovations (Hagedoorn and Schakenraad, 1993), and resulting economic performance on firm level (Ernst, 2001). These signals are also useful and recognized by investors (Hirschey and Richardson, 2004), making them more likely to provide firms with external capital (Hall and Harhoff, 2012). Yet, it has widely been recognized that the technological as well as economic significance of patents varies broadly (Basberg, 1987). Consequently, the *ex-ante* identification of potential high value and impact is of high relevance for firms, investors, and policy makers alike. Besides guiding the allocation of investments and internal resources, it might enable "nowcasting" and "placecasting" of the quality of inventive and innovative activity (consider Andrews et al., 2017; Fazio et al., 2016; Guzman and Stern, 2015, 2017, for an application in entrepreneurship). However, by definition patents of abnormally high value and impact are rare in nature. Together with the broad availability of structured patent data via providers such as PATSTAT and the OECD, this makes the presented setting an useful and informative case for a predictive modeling exercise.

---

[9]Bootstrapping is a technique most applied econometricians are well-acquainted with, yes used for a slightly different purpose. In econometrics, bootstrapping represents a powerful way to circumvent problems arising out of selection bias and other sampling issues, where the regression on several subsamples is used to adjust the standard errors of the estimates.

### 3.2 Data

For this exercise, we draw from the patent database provided by PATSTAT. To keep the data volume moderate and the content somewhat homogeneous, we here limit ourselves to patents granted at the USTPO and EPO in the 1980-2016 period, leading to a number of roughly 6.5 million patents. While this number appears already large compared to other datasets commonly used by applied econometricians in the field of entrepreneurship and innovation studies, according to ML standards it can still be considered as small, both in terms of the number of observation as well as available variables. While offering reasonable analytic depth, such amounts of data can still be conveniently processed with standard in-memory workflows on personal computers.[10]

We classify high impact patents following (Ahuja and Lampert, 2001) as the patents within a given cohort receiving the most citations by other patents within the following 5 year window. Originally, such "breakthrough patents" are defined as the ones in the top 1% of the distribution. For this exercise, we also create another outcome, indicating the patent to be in the top 50% of the distribution, indicating above-average successful but not not necessarily breakthrough patents.

For the sake of simplicity and reconstructability, we in the following models mostly use features either directly contained in PATSTAT, or easily derived from it. In detail, we create a selection of *ex-ante* patent novelty and quality indicators,[11] summarized in table 1. The only data aexternal to PATSTAT data we used are the temporal technological similarity indicator proposed by Hain (2018).

---

[10] This is often not the case for typical ML problems, drawing from large numbers of observations and/or a large set of variables. Here, distributed or cloud-based workflows become necessary. We discuss the arising challenges elsewhere (e.g., Hain and Jurowetzki, ming).

[11] For a recent and exhaustive review on patent quality measures, including all used in this exercise, consider Squicciarini et al. (2013)

Table 1: Descriptive Statistics: USTPO Patents 2010-2015

| Feature | N | Mean | St. Dev. | Min | Max | Description |
|---|---|---|---|---|---|---|
| breakthrough | 6,571,931 | 0.006 | 0.074 | 0 | 1 | Top 1%-cited patent in annual cohort |
| breakthrough50 | 6,571,931 | 0.175 | 0.380 | 0 | 1 | Top 50%-cited patent in annual cohort |
| sim.past | 6,571,931 | 0.088 | 0.185 | 0 | 1 | Technological similarity to past (Hain, 2018) |
| sim.present | 6,571,931 | 0.153 | 0.262 | 0 | 1 | Technological similarity to present (Hain, 2018) |
| many_field | 6,571,931 | 0.398 | 0.489 | 0 | 1 | Multiple IPC classes (Lerner, 1994) |
| patent_scope | 6,571,931 | 1.854 | 1.162 | 1 | 31 | Number of IpC classes |
| family_size | 6,571,931 | 4.251 | 3.906 | 1 | 57 | Size of the patent family (Harhoff et al., 2003) |
| bwd_cits | 6,571,931 | 15.150 | 25.640 | 0 | 4,756 | Backward citations (Harhoff et al., 2003) |
| npl_cits | 6,571,931 | 3.328 | 12.690 | 0 | 1,592 | NPL backward citations (?) |
| claims_bwd | 6,571,931 | 1.673 | 3.378 | 0 | 405 | Backward claims |
| originality | 6,571,931 | 0.707 | 0.248 | 0 | 1 | *originality* index (Trajtenberg et al., 1997) |
| radicalness | 6,571,931 | 0.382 | 0.288 | 0 | 1 | *radicalness* index (Shane, 2001) |
| nb_applicants | 6,571,931 | 1.849 | 1.705 | 0 | 77 | Number of applicants |
| nb_inventors | 6,571,931 | 2.666 | 1.925 | 0 | 99 | Number of Inventors |
| patent_scope.diff | 6,571,931 | 0.008 | 1.091 | -2.806 | 29.130 | $\Delta$ patent–cohort scope |
| bwd_cits.diff | 6,571,931 | 0.222 | 24.560 | -42.050 | 4,732.000 | $\Delta$ patent–cohort backwards citations |
| npl_cits.diff | 6,571,931 | 0.112 | 12.100 | -30.510 | 1,579.000 | $\Delta$ patent–cohort nlp citations |
| family_size.diff | 6,571,931 | 0.031 | 3.536 | -11.090 | 50.090 | $\Delta$ patent–cohort family size |
| originality.diff | 6,571,931 | -0.029 | 0.242 | -0.911 | 0.431 | $\Delta$ patent–cohort originality |
| radicalness.diff | 6,571,931 | -0.018 | 0.277 | -0.751 | 0.808 | $\Delta$ patent–cohort radicalness |
| sim.past.diff | 6,571,931 | -0.000 | 0.180 | -0.247 | 0.990 | $\Delta$ patent–cohort similarity to past |
| sim.present.diff | 6,571,931 | 0.000 | 0.260 | -0.315 | 0.942 | $\Delta$ patent–cohort similarity present |

Notice that the breakthrough features are calculated based on the distribution of patents that receive citations. Since a growing number of patents never get cited, the percentage of patents that fall within the top-n% appears less than expected. Notice also that we abstain of including a lot of categorical features, which would traditionally be included in a causal econometric exercise, for example dummies for the patent's application year, the inventor and applicant firm. Again, since we aim at building a predictive model that fits well on new data. Obviously, such features would lead to overfitting, and reduce its performance when predicting up to now unobserved firms, inventors, and years. We include dummy features for the technological field, though, which is a somewhat more static classification. However, since the importance of technological fields also change over time, such a model would be in need of retraining as time passes by and the importance of technological fields shift.

### 3.3 First Data Exploration

The ML toolbox around predictive modeling is rich and diverse, and the variety of available techniques in many cases can be tuned along a set of parameters, depending on the structure of data and problem at hand. Therefore, numerical as well as visual data inspection becomes an integral part of the model building an tuning process.

First, for our set of features to be useful for a classification problem, it is useful (and for the autoencoder model introduced later necessary) they indeed display a different distribution conditional to the classes we aim at predicting. In figure 4, we plot this conditional distribution for all candidate features for our outcome classes of `breakthrough` and `breakthrough50`, where we indeed observe such differences.

Figure 4: Conditional distribution of Predictors

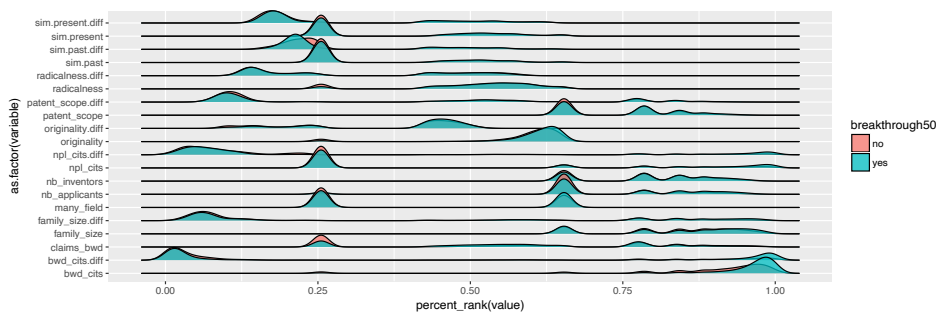

(a) Conditional to *breakthrough50* ($\geq 50\%$ forward citations in cohort)

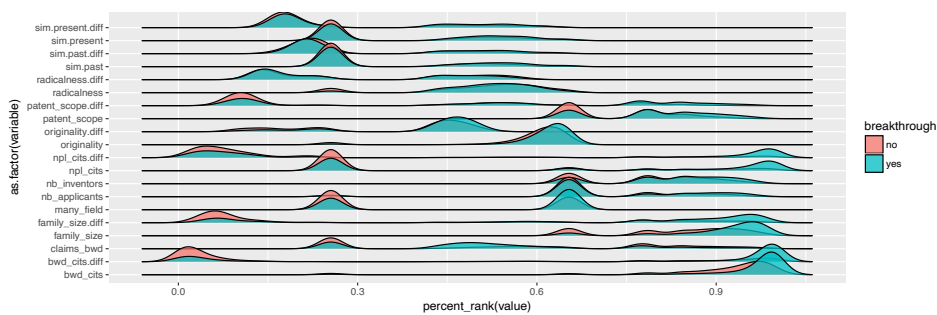

(b) Conditional to *breakthrough01* ($\geq 99\%$ forward citations in cohort)

## 3.4 Preprocessing

Again, as an remainder, for a predictive modeling exercise we are *per se* not in need of producing causal, robust, or even interpretative parameter estimates. Consequently, we enjoy a higher degree of flexibility in the way how we select, construct, and transform model features. First, moderate amounts of missing feature values are commonly imputed, while observations with missing outcomes are dropped. Second, various kind of "feature scaling" techniques are usually performed in the preprocessing stage, where feature values are normalized in order to increase accuracy as well as computational efficiency of predictive models. The selection of appropriate scaling techniques again depends on the properties of the data and model at hand, where popular approaches are minMax rescaling ($x' = \frac{x-\bar{x}}{\sigma}$), mean normalization ($x' = \frac{x-\text{mean}(x)}{\max(x)-\min(x)}$), standardization ($x' = \frac{x-\bar{x}}{\sigma}$), dimensionality reduction of the feature space with a principal component analysis (PCA), and binary recoding to "one-hot-encodings". In this case, we normalize all continuous features to $\mu = 0$, $\sigma = 1$, and categorical features to one-hot-encoding.

Before we do so, we split our data in the test sample we will use for the model and hyperparameter tuning phase (75%), and a test sample, which we will only use once for the final evaluation of the model (25%). It is important to do this split before the preprocessing, since otherwise a common feature scaling with the test sample could contaminate our training sample.

### 3.5 Model Setup and Tuning

After exploring and preprocessing our data, we now select and tune a set of models to predict high impact patents, where we start with the outcome *breakthrough3*, indicating a patent to be among the 50% most cited patents in a given year cohort. This classification problem calls for a class of models able to predict categorical outcomes. While the space of candidate models is vast, we limit ourselves to the demonstration of popular and commonly well performing model classes, namely the traditional logit, elastic nets, boosted classification trees, and random forests. Most of these models include tunable hyperparameter, leading to varying model performance. Given the data at hand, we aim at identifying the best combination of hyperparameter values for every model before we evaluate their final performance and select the best model for our classification problem.

We do so via a hyperparameter "grid search" and repeated 5-fold crossvalidation, For every hyperparameter, we define a sequence of possible values. In case of multiple hyperparameters, we create a tune grid, a matrix containing a cell for every unique combination of hyperparameter values. Then we perform the following steps:[12]

1. Partition the training data into 5 equally sized folds.

2. Fit a model with a specific hyperparameter combination separate on fold 1-4, evaluate its performance by predicting the outcome of fold 5.

3. Repeat the process up to now 5 times.

4. Calculate the average performance of a hyperparameter combination.

5. Repeat the process up to now for every hyperparameter combination.

6. Select the hyperparameter combination with the best average model performance.

7. Fit the final model with optimal hyperparameters on the full training data.

It is easy to see that this exhaustive tuning process results in a large amount of models to run, of which some might be quite computationally intensive. To reduce the time spent on tuning, we here separate hyperparameter tuning and fitting the final model, where the tuning is done on only a subset of 10% of the training data, and only the fit of the final model on the full training data.

---

[12] While the described process appears rather tedious by hand, specialized ML packages such as `caret` in R provide efficient workflows to automate the creation of folds as well as hyperparamether grid search.

### 3.5.1 Logit

The class of logit regressions for binary outcomes is well known and applied in econometrics and ML alike, and will serve as a baseline for more complex models to follow. In its relatively simple and rigid functional form, there are no tunable parameters. Taking this functional form as given, then minimizing the out-of-sample loss function $L(\hat{y}, y)$ becomes a question of (i.) how many variables, and (ii.) which variables to include. Such problems of *variable selection* are well known to econometricians, which use them mainly for the selection of control variables, including stepwise regressions (one-by-one adding control variables with the highest impact on our $\bar{R}^2$), partial least squares (PLS), different information criterion (e.g., Aikon: AIC, Bayesian: BIC), to only name a few.[13] While in reality there might be incentives to create a sparse model, we for this exercise use all variables available.

### 3.5.2 Elastic Nets

We proceed a second parametric approach, a class of estimators for penalized linear regression models that lately also became popular among econometricians, *elastic nets*. Generally, the functional form is identical to a generalized linear model, with a small addition. Our $\beta$ parameters are weighted by an adittional parameter $\lambda$, which penalizes the coefficient by its contribution to the models loss in the form of:

$$\lambda \sum_{p=1}^{P} [1 - \alpha |\beta_p| + \alpha |\beta_p|^2] \tag{2}$$

Of this general formulation, we know two popular cases. When $\alpha = 1$, we are left with the quadratic term, leading to a *ridge regression*. If $\alpha = 0$, we are left with $|\beta_i|$, turning it to a lately among econometricians very popular "Least Absolute Shrinkage and Selection Operator" (LASSO) regression. Obviously, when $\lambda = 0$, the whole term vanishes, and we are again left with a generalized linear model[14] Consequently, the model has two tunable parameters, $\alpha$ and $\lambda$, over which we perform a grid search., illustrated in figure 5.

While for low $\alpha$ values the model performance turns out to be somewhat insensitive to changes in $\lambda$, with increasing $\alpha$ values, 1 leads to sharply decreasing model performance. With a slight margin, the pest performing hyperparameter configuration resembles a LASSO ($= 1, \alpha = 0$).

---

[13] For an exhaustive overview on model and variable selection algorithms consider Castle et al. (2009).

[14] For an exhaustive discussion on the use of LASSO, consider Belloni et al. (2014). Elastic nets are integrated, among others, in the R package `Glmnet`, and Python's`scikit-learn`.

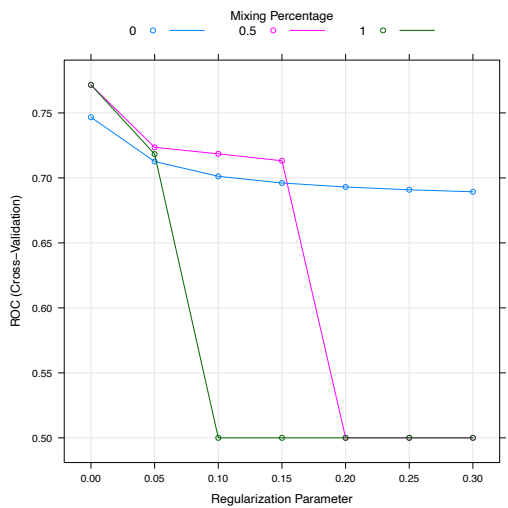

Figure 5: Hyper-Parameter Tuning Elastic Nets

### 3.5.3 Classification Tree

Next, this time following a non-parametric approach, we fit a *classification and regression trees* (CART, in business application also known as *decision trees*). [15] The rich class of classification trees is characterized by a flexible functional form able to fit complex relationships between predictors and outcomes, yet is can be illustrated in an accessible way. They appear to show their benefits over traditional logistic regression approaches mostly in settings where we have a large sample size (Perlich et al., 2003), and where the underlying relationships are really non-linear (Friedman and Popescu, 2008). The general idea behind this approach is to step-wise identify feature explaining the highest variance of outcomes. This can be done in various ways, but in principle you aim to at every step use some criterion to identify the most influential feature *X* of the model (e.g., the lowest *p* value), and then another criterion (e.g., lowest $\chi^2$ value) to determine a cutoff value of this feature. Then, the sample is split according to this cutoff. This is repeated for every subsample, leading to a tree-like decision structure, which eventually ends at a terminal node (a *leaf*), which in the optimal case contains only or mostly observation of one class. While simple and powerful, classification trees are prone to overfitting, when left to grow unconstrained, since this procedure can be repeated until every observation ends up in an own leaf, characterized by an unique configuration of features. In practice, a tree's complexity can be constrained with a number of potential hyperparameters, including a limit the maximum depth, or criteria if a split is accepted or the node becomes a terminal one (e.g., certain $p-value$, certain improvement in the predictive performance, or a minimum number of observations falling in a split).

---

[15] There are quite many packages dealing with different implementations of regression trees in common data science environments, such as `rpart, tree, party` for R, and again the machine learning allrounder `scikit-learn` in Python. For a more exhaustive introduction to CART models, consider Strobl et al. (2009)

In this exercise, we fit a classification tree via a "Recursive Partitioning" implemented in the `rpart` package in R, cf. Therneau et al., 1997. The resulting tree structure can be inspected in figure **??**.

Figure 6: Structure of the Decision Tree

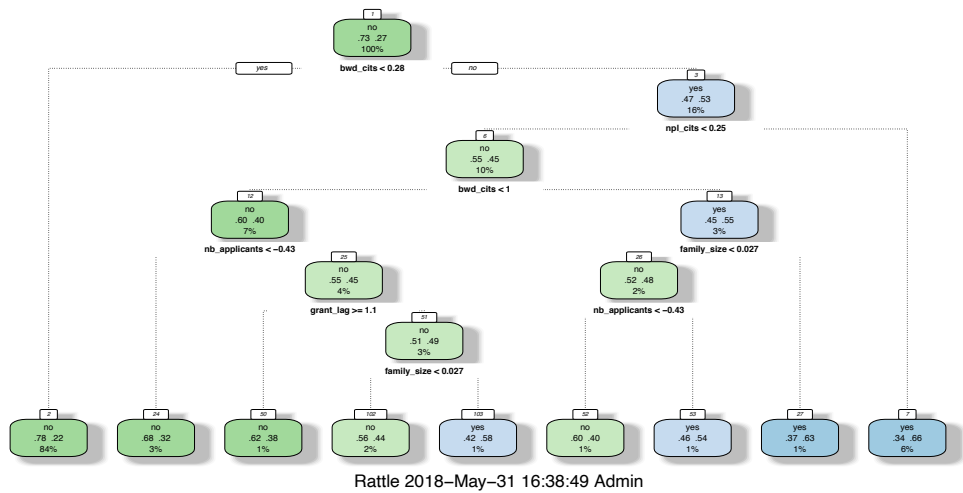

Here we are able to restrict the complexity via a hyperparameter $\alpha$. This parameter represents the complexity costs of every split, and allows further splits only if it leads to an decrease in model loss below this threshold. Figure 7 plots the result of the hyperparameter tuning of $\alpha$.

We directly see that in this case, increasing complexity costs lead to decreasing model performance. Such results are somewhat typical for large datasets, where high complexity costs prevent the tree to fully exploit the richness of information. Therefore, we settle for a minimal $\alpha$ of 0.001.

### 3.5.4 Random Forest

Finally, we fit another class of models which has gained popularity in the last decade, and proven to be a powerful and versatile prediction technique which performs well in almost every setting, a random forest. As a continuation of tree-based classification methods, random forests aim at reducing overfitting by introducing randomness via bootstrapping, boosting, and ensemble techniques. The idea here is to create an "ensemble of classification trees", all grown out of a different bootstrap sample. These trees are typically not pruned or otherwise restricted in complexity, but instead, a random selection of features is chosen to determine the split at the next decision nodes.[16] Having grown a "forest of trees", every tree performs a prediction, and the final model prediction is formed by a "majority vote" of all trees. The idea is close to the Monte-Carlo approach, assuming a large population of weak predictions injected by randomness leads to overall stronger results than one strong and potentially overfitted prediction. However, the robustness of this model class comes with a price. First, the large amount of models to be fitted is computationally rather intensive, which becomes painfully visible when working with large datasets. Further, the predictions made by a random forest are more opaque than the ones provided by the other model classes used in this example. While the logit and elastic net delivers easily interpretable parameter estimates and the classification tree provides a relatively intuitive graphical representation of the classification process, there exists no way to represent the functional form and internal process of a classification carried out by a random forest in a way suitable for human annotation.

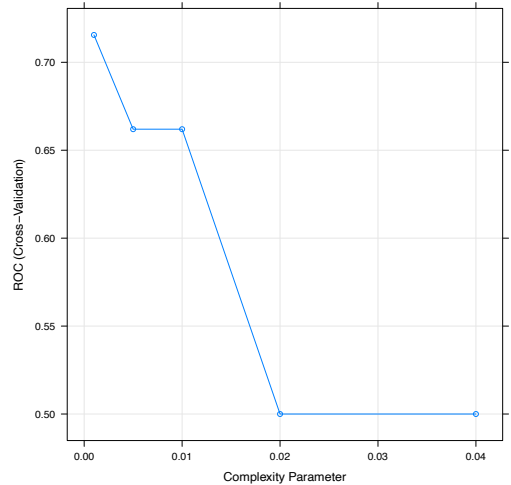

Figure 7: Hyper-Parameter Tuning Classification Tree

In this case, we draw from a number of tunable hyperparameters. First, we tune the number of randomly selected features which are available candidates for every split on a range $[1, k-1]$, where lower values introduce a higher level of randomness to every split. Our second hyperparameter is the minimal number of observations which have to fall in every split, where lower numbers increase the potential precision of splits, but also the risk of overfitting. Finally, we also use the general splitrule as an hyperparameter, where the choice is between i.) a traditional split according to a the optimization of the gini coefficient of the distribution of classes in every split, and ii.) according to the "Extremely randomized trees" (ExtraTree) procedure by Geurts et al. (2006), where adittional randomness is introduced to the selection of splitpoints.

---

[16]Indeed, it is worth mentioning here that many model tuning techniques are based on the idea that adding randomness to the prediction process – somewhat counter-intuitively – increases the robustness and out-of-sample prediction performance of the model.

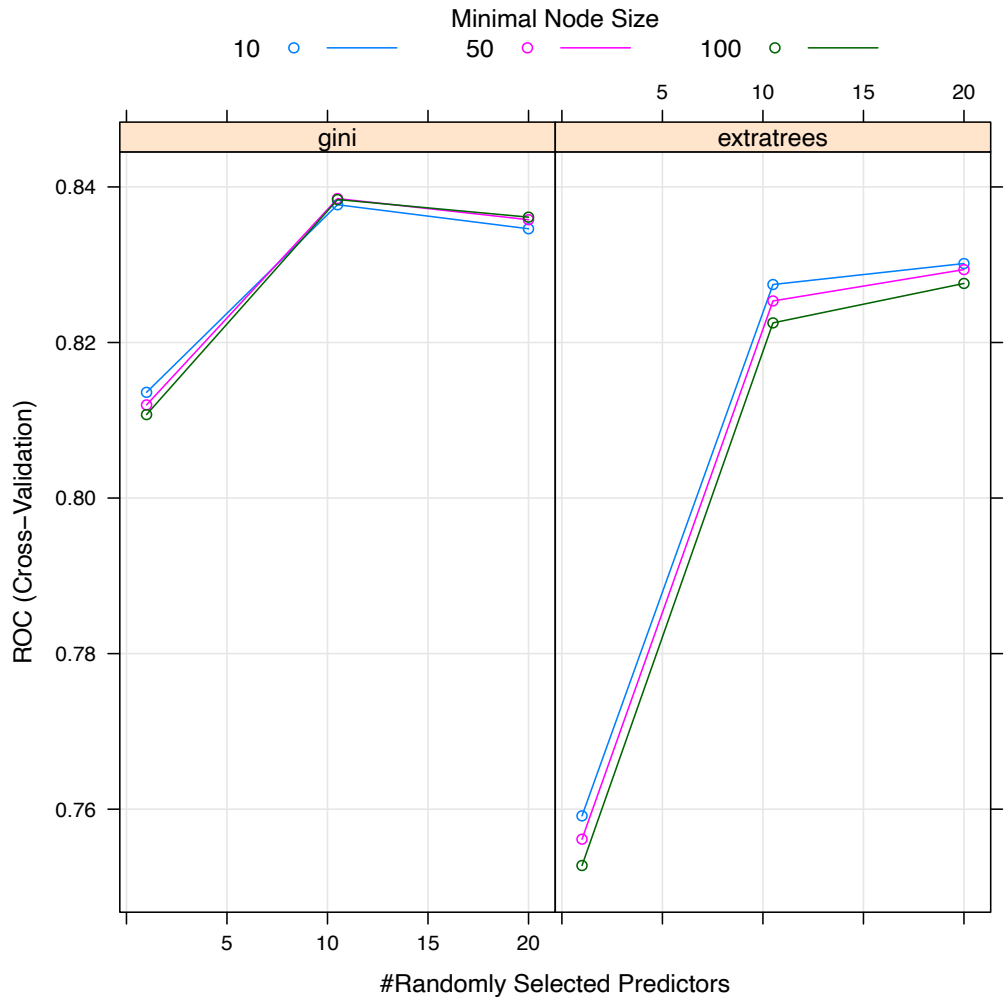

Figure 8: Hyper-Parameter Tuning Random Forest

In figure 8 we see that number of randomly selected features per split of roughly half (22) of all available features in all cases maximizes model performance. Same goes for a high minimal number of observations (100) per split. Finally, the ExtraTree procedure first underperforms at a a minimal amount of randomly selected features, but outperforms the traditional gini-based splitrule when the number of available features increases. Such results are typical for large samples, where a high amount of injected randomness tends to make model predictions more robust.

### 3.5.5 Explainability

In the exercise above we demonstrate that richer model classes with a flexible functional form indeed enable us to better capture complex non-linear relationships and enable us to tackle hard prediction problems more efficiently that traditional methods and techniques from causal modeling, which are usually applied by econometricians. First, even in parametric approaches, feature effects in predictive modeling are explicitly non-causal.[17] This holds true for most machine learning approaches and represents a danger for econometricians using them blindly. Again, while an adequately tuned machine learning model may deliver very accurate estimates, it is misleading to believe that a model designed and optimized for predicting $\hat{y}$ *perse* also produces $\beta$'s with the statistical properties we usually associate with them in econometric models.

Second, with increasing model complexity, the prediction process becomes more opaque, and the isolated (non-causal) effect of features on the outcome becomes harder to capture. While the feature effect of the logit and elastic net can be interpreted in a familiar and straightforward way as a traditional regression table, already in classification tree models (see figure **??**) we do not get constant *ceteris paribus* parameter estimates. However, the simple tree structure still provides some insights into the process that leads to the prediction. The predictions of a forest consisting of thousands of such trees in a random forest obviously cannot be interpreted anymore in a meaningful way.

Some model classes have developed own metrics of variable impact as we depict for our example in table 10. However, they are not for all model classes available, and sometimes hard to compare across models.In this cases, the most straightforward war to get an intuition of feature importance across models is to calculate the correlation between the features and predicted outcome, as we did in figure . Again, this gives us some intuition on the relative influence of a feature, but tells us nothing about any local prediction process.

From investigating the relative variable importance, we gain a set of insights. First, we see quite some difference in relative variable importance across models. While backward citations across all models appears to be a moderate or good predictor, technology fields are assigned as highly predictive in the elastic net, but way less in the random forest, that ranks all other features higher. Furthermore, the extend to which the models draw from the available features differs. While the elastic net draws more evenly from a large set of features, in the classification tree only 8 are integrated. That again also reminds us that features importance, albeit informative, cannot be interpreted as a causal effect.

---

[17] Just to give an example, Mullainathan and Spiess (2017) demonstrate how a LASSO might select very different features in every fold.

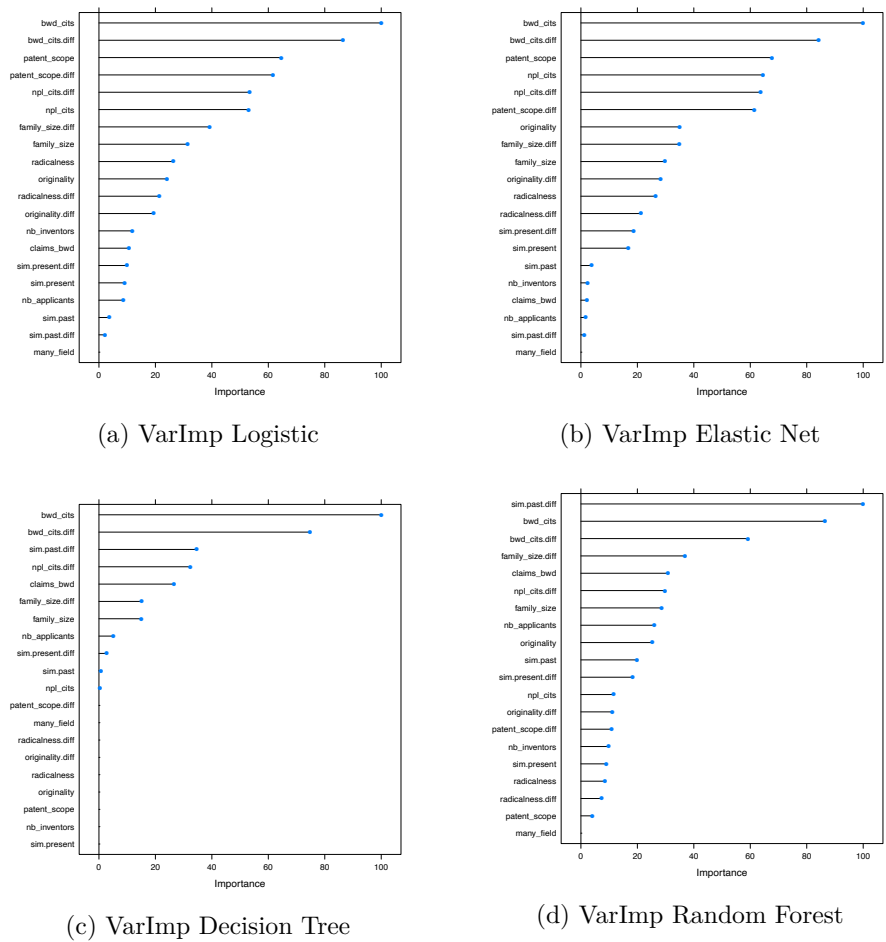

Figure 9: Variable importance of final models

(a) VarImp Logistic

(b) VarImp Elastic Net

(c) VarImp Decision Tree

(d) VarImp Random Forest

### 3.5.6 Final Evaluation

After identifying the optimal hyperparameters for every model class, we now fit the final prediction models on the whole training data accordingly. As a final step, we evaluate the performance of the models by investigating its performance on a holdout sample, consisting of 25% of the original data, which was from the start set aside and never inspected, or used for any model fitting. Figure 10 displays the results of the final model's prediction on the holdout sample by providing a confusion matrix as well as the ROC curve to the corresponding models, while table **??** provides a summary of standard evaluation metrics of predictive models.

We see that the logit and elastic net in this case leads to almost identical results across a variety of performance measures. Surprisingly, the classification tree in this case is the worst performing model, despite it's more complex functional form. However, earlier we saw that the classification tree takes only a small subset of variables into account, hinting at a to restrictive choice of the complexity parameter $\alpha$ during the hyperparameter tuning phase. This is also indicated by the kinked ROC curve in 10, indicating that after exploiting the effect of a few parameters the model has little to contribute. Across all measures, the random forest, as expected, performs best. Here we see how randomization and averaging over a large number of repetition indeed overcomes many issues of the classification tree. While the overall accuracy (ratio of correctly classified to all observations) only increases modestly, the low occurrence of positive outcomes makes this measure only partially informative in this case. However, the sensitivity (ratio correctly classified positives to all positive observation), increases quite visibly compared to other models. This also highlights issues of traditional methods often occurring when facing unbalanced outcome states.

Table 2: Final Model Evaluation with Test Sample

| names       | Logit | ElasticNet | ClassTree | RandForest |
|-------------|-------|------------|-----------|------------|
| Accuracy    | 0.758 | 0.758      | 0.754     | 0.77       |
| Kappa       | 0.228 | 0.226      | 0.217     | 0.299      |
| Sensitivity | 0.222 | 0.219      | 0.219     | 0.304      |
| Specificity | 0.959 | 0.96       | 0.953     | 0.944      |
| AUC         | 0.73  | 0.73       | 0.607     | 0.758      |

Up to now we demonstrated a predictive modelling workflow using traditional machine learning techniques, which performs reasonably well for the exercise at hand. While the prediction of successful patents can be seen as analytically non-trivial, we are able to achieve an acceptable level of accuracy by only using features that can be computed directly from the PATSTAT data with little effort. With a more complex feature generation procedure (eg., matching with further datasources to include inventor or applicant characteristics), this performance could likely be improved further. However, be reminded that up to now we only fitted models to predict *breaktrough3*, the outcome indicating that the patent is in the top 50% of received citations within its cohort (ca. 37% of cases). We repeated this exercise for the outcome *breakthrough* (top 1%, 0.6% of cases), where we get less optimistic results. For our initial target outcome, the rare *breakthrough*, all models are of no help whatsoever and predict 100% non-breakthroughs, except of the random forest, which here predicts marginally better than a coin toss.

### 3.6 Outlook: Deep Learning and Rare Event Prediction

Up to now, we demonstrated how various more traditional model classes can be applied in a predictictive modelling context at the example of a classification problem

Figure 10: ROC curves of final models

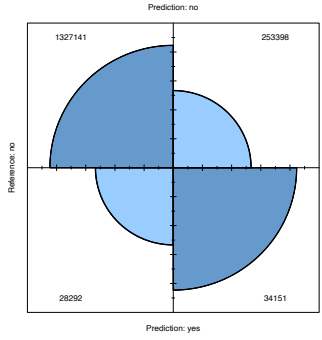

(a) ConvMat Logistic

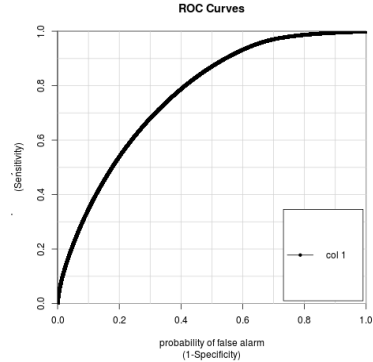

(b) ROC Logit

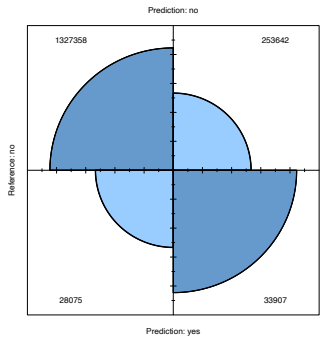

(c) ConvMat Elastic Net

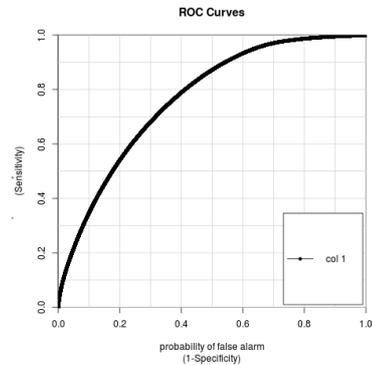

(d) ROC Elastic Net

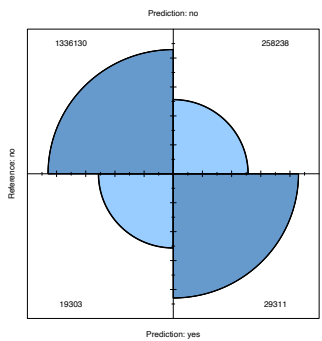

(e) ConvMat Classification Tree

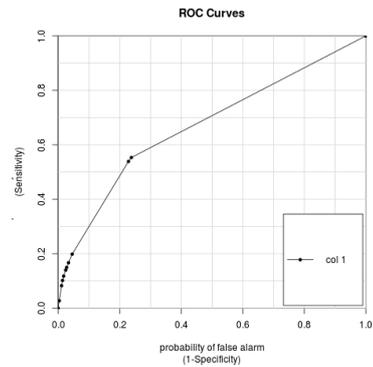

(f) ROC Classification Tree

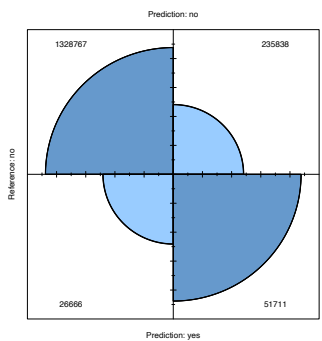

(g) ConvMat Random Forest

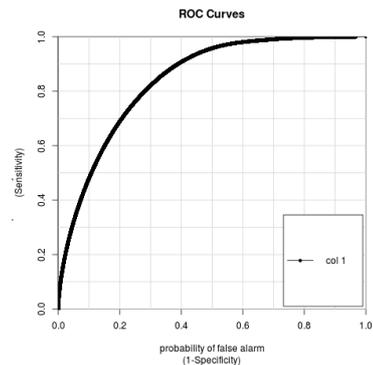

(h) ROC Random Forest

with approximately 15% of positive cases of interest. But what if we were interested in the "breakthrough" patents? The really rare events that receive large amounts of citations? This is a situation where more traditional approaches fail. In such cases the dataset is heavily unbalanced. Most supervised machine learning approaches are sensitive to such scenarios and a common approach has been to undersample the data. More recently deep learning approaches to such problems have become popular, due to their efficiency, ability to handle large amount of data in a sequential manner (data can be streamed line by line) and flexibility that allows to work with different types of inputs (cross-sectional, sequential, text, images etc.). In the following we will start by repeating the exercise from the previous section using a relatively simple neural network architecture. Then, we will apply the same architecture to the breaktrough-identification problem. Lastly, we will use a deep autoencoder and an anomaly detection approach to identify the most rare breaktrough patents. Before that, we will, however, provide a short, more general introduction to artificial neural networks and deep learning.

### 3.6.1 Introduction to Artificial Neural Networks and Deep Learning

Regression trees might still be familiar to some applied statisticians and quantitative researchers. Now we would like to introduce to another class of models which due to current breakthroughs delivered unprecedented prediction performance on large high-dimensional data and thus enjoys a lot of popularity: Neural networks. Connecting to the former narrative, neural networks represent *regression trees on steroids*, which are flexible enough to – given enough data – fit every functional form and thereby theoretically can produce optimal predictions to every well-defined problem.

While early ideas about artificial neural networks (ANNs) were already developed in the 1950s and 60s by among others Frank Rosenblatt 1958 and the formal logic of neural calculation described by McCulloch and Pitts (1943), it took several decades for this type of biology-inspired models to see a renaissance in the recent few years.[18]

This revival can be attributed to three reasons: (i) New training techniques, (ii) the availability of large training datasets, and (iii) hardware development, particularly the identification of graphical processing units (GPUs) – normally used, as the name suggests, for complex graphics rendering tasks in PCs and video game consoles – as extremely well suited for modeling neural networks (LeCun et al., 2015).

---

[18]It has to be stressed that even though neural networks are indeed inspired by the most basic concept of how a brain works, they are by no means mysterious artificial brains. The analogy goes as far as the abstraction that a couple of neurons that are interconnected in some architecture. The neuron is represented as some sigmoid function (somewhat like a logistic regression) which decides based on the inputs received if it should get activated and send a signal to connected neurons, which might again trigger their activation. Having that said, calling a neural network an artificial brain is somewhat like calling a paper-plane an artificial bird.

To understand the neural network approach to modeling, it is essential to get a basic grasp of two main concepts. First, the logic behind the functioning of single neurons[19], and second the architecture and sequential processes happening within ANNs.

A single neuron receives the inputs $x_1,x_2,...,x_n$ with weights $w_1,w_2,...,w_n$ that are passed to it through "synapses" from previous layer neurons (i.e. the input layer). Given these inputs the neuron will "fire" and produce an output $y$ passing it on to the next layer, which can be a hidden layer or the output layer. In praxis, the inputs can be equated to standardized or normalized independent variables in a regression function. The weights play a crucial role, as they decide about the strength with which signals are passed along in the network. As the network learns, the initially randomly assigned weights are continuously adjusted. As the neuron receives the inputs, it first calculates a weighted sum of $w_i x_i$ and then applies an activation function $\phi$.

$$\phi(\sum_{i=1}^{m} w_i x_i) \tag{3}$$

Depending on the activation function the signal is passed on or not.

Figure 11: Illustration of a neuron

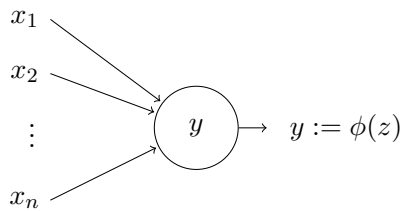

Figure 12 represents an artificial neural network with three layers: One input layer with four neurons, one fully connected hidden layer with five neurons and one output layer with a single neuron. As the model is trained for each observation inputs are passed on from the input layer into the neurons of the hidden layer and processed as described above. This step is repeated and an output value $\hat{y}$ is calculated. This process is called forward propagation. Comparing this value with the actual value $y$ (i.e. our dependent variable for the particular observation) allows to calculate a *cost function* e.g. $C = \frac{1}{2}(\hat{y} - y)^2$. From here on *backpropagation*[20] is used to update the weights. The network is trained as these processes are repeated for all observations in the dataset.

Artificial neural networks have many interesting properties that let them stand out from more traditional models and make them appealing when approaching complex

---

[19] for the sake of simplicity here we will not distinguish between the simple perceptron model, sigmoid neurons or the recently more commonly used rectified linear neurons (Glorot et al., 2011)

[20] This complex algorithm adjusts simultaneously all weight in the network, considering the individual contribution of the neuron to the error.

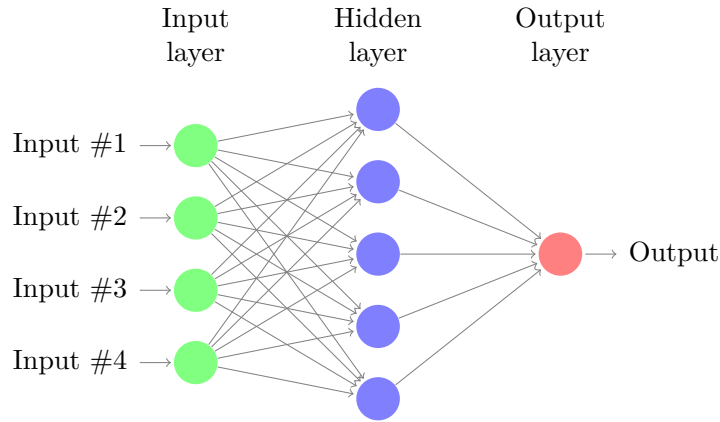
Figure 12: Illustration of a neural network

pattern discovery tasks, confronting nonlinearity but most importantly dealing with large amounts of data in terms of the number of observations and the number of inputs. These properties, coupled with the recent developments in hardware and data availability, led to a rapid spread and development of artificial nets in the 2010s. Today, a variety of architectures has evolved and is used for a large number of complex tasks such as speech recognition (Recurrent neural networks: RNNs and Long Short Term Memory: LSTMs), computer vision (CNNs and Capsule Networks, proposed in late October 2017) and as backbones in artificial intelligence applications. They are used not only because they can approach challenges where other classes of models struggle technically but rather due to their performance.

Despite the numerous advantages of artificial neural nets they are yet rarely seen applied in non-technical research fields. Here CNNs may be so far the most often used type, where its properties were employed to generate estimates from large image datasets Gebru et al. (e.g. 2017). The simplest architecture of a CNN puts several convolutional, and pooling layers in front of an ANN. This allows transforming images, which are technically two-dimensional matrices, into long vectors, while preserving the information that describes the characteristic features of the image.

The predictive performance of neural nets stands in stark contrast to the explainability of these models, meaning that a trained neural net is more or less a black box, which produces great predictions but does not allow to make causal inference. In addition, this leads to asking: What is the reason the model produced this or that prediction. This becomes particularly important when such models are deployed for instance in diagnostics or other fields to support decision making. There are several attempts to address this problem under the heading of "explainable AI" (e.g. Ribeiro et al., 2016).

### 3.6.2 Application: Simple feed forward neural network for predicting successful and breakthrough patents

Before approaching a complex neural architecture in the following section, here we use a relatively simple feed forward neural network to first tackle the problem, presented in earlier sections. The network is composed of 3 dense layers of 22, 20, 15 neurons respectivly and an output layer. In addition two dropout layers with 0.3 and 0.2 rates have been added for regularization. We used gridsearch and k-fold cross-validation to determine optimal hypermarameter settings. Such a setup represents a relatively standard architecture and no further variation has been explored regarding adding and removing of further neurons or layers.

### 3.6.3 Application: Prediction of breakthrough patents as anomaly detection with Stacked Autoencoder

An autoencoder is a neural network architecture that is used to learn a representation (encoding) for a set of data. Autoencoders are referred to as "self-suprevised" models, as their inputs are usually identical to their outputs (Hinton and Salakhutdinov, 2006).

$$f_{W,b}(x) \approx x \tag{4}$$

Figure 13: Illustration of an autoencoder architecture

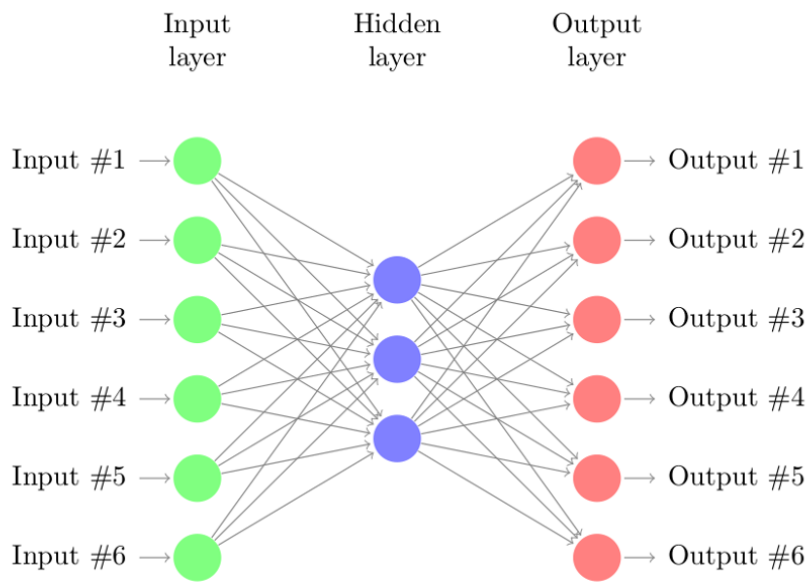

The typical use case is dimensionality reduction: Training such models to reconstruct some inputs from a low-dimensional representation in the latent hidden layer as shown in Figure 13 makes them very powerful general dimensionality reduction tools. In comparison to techniques such as PCA, autoencoders capture all kinds of non-linearities and interactions. The encoder part of such a model can be used separately to create some low-dimensional representations of inputs that consequently can be for instance clustered or used to identify similar inputs, which is used in state-of-the-art recommender systems (Sedhain et al., 2015). In combination with recurrent or convolutional layers inputs can be sequences or images, respectively (Tian et al., 2014). Sequence to Sequence (Seq2seq) models with recurrent layers on both ends are increasingly used for neural machine translation and contributed to great improvements in the field (Sutskever et al., 2014). Such models are trained by using phrases in the source language as inputs and the same in the target language as outputs. The hidden layer of such models is called "thought" vector, as it incodes a universal representation meaning representation of the input that the model can translate into another language. Another recent application of autoencoders has been anomaly detection (Sakurada and Yairi, 2014). The idea is very simple: Given the vast amount of "normal" cases, it is easy to train the autoencoder to "reconstruct" the *status quo*. This information will constitute all what the autoencoder is exposed to, and therefore we would expect that non-normal inputs would confuse the autoencoder. Two approaches to capture such confusion are proposed[21]: We could now (a) extract the latent layer and cluster it, reduce dimensionality with for example PCA and measure the distance of the known outlyers to the centeroids of our the clusters (Sakurada and Yairi, 2014). (b) An even easier approach is to look at the behaviour of the autoencoder more directly, particularly through the reconstruction error – e.g. Mean Square Error or Eucledian Distance, Equation 5 and 6 respectively (An and Cho, 2015).

$$L(x, x') = \| x - x' \|^2 \tag{5}$$

$$L(x, x') = \| x - x' \|_2 \tag{6}$$

In a nutshell: An autoencoder that is really good at reconstructing "boring normality" should experience considerable difficulties when trying to reconstruct anomaly events.

The latter is the approach that we take to detect the the 0.6% breakthrough patents in our dataset that more traditional models would most likely oversee. We use the same data as above, however removing the categorical variable `technology field`. All other variables are normalised.

---

[21]For some overview on other methods using similar logic, consider: Shyu et al. (2003); Wang (2005); Zhou and Lang (2003)

Figure 14: Autoencoder Model Architecture Summary

```
Layer (type)                    Output Shape                  Param #
=================================================================
input_2 (InputLayer)            (None, 11)                    0
_________________________________________________________________
dense_5 (Dense)                 (None, 9)                     108
_________________________________________________________________
dense_6 (Dense)                 (None, 4)                     40
_________________________________________________________________
dense_7 (Dense)                 (None, 4)                     20
_________________________________________________________________
dense_8 (Dense)                 (None, 11)                    55
=================================================================
Total params: 223
Trainable params: 223
Non-trainable params: 0
_________________________________________________________________
```

Figure 15: ROC-AUC for the Stacked Autoencoder

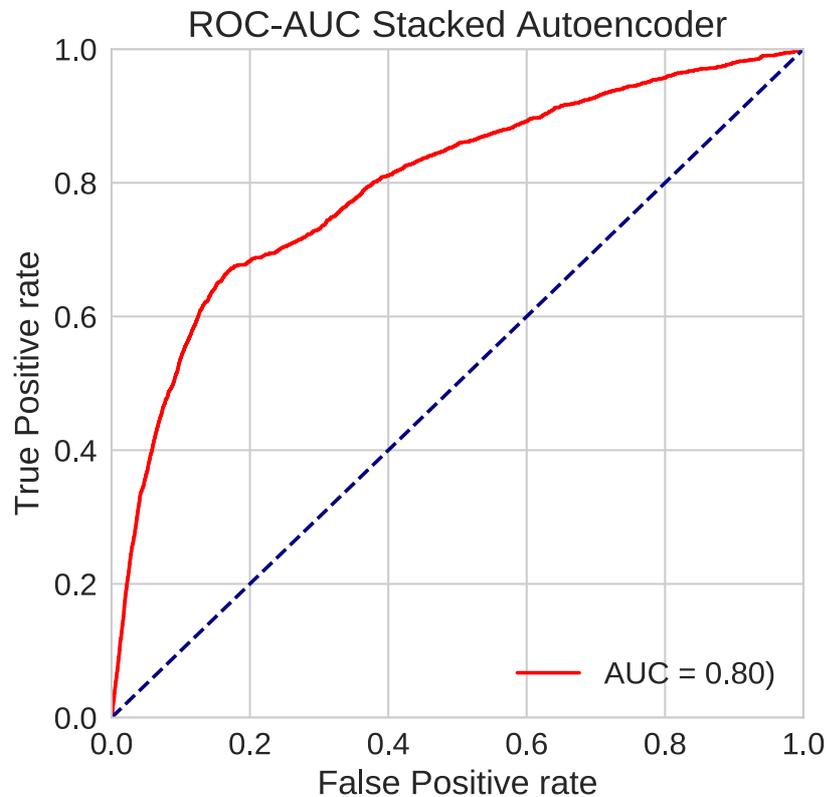

We train a stacked autoencoder model using Tensorflow-Keras in Python.[22] Neural network models, and particularly autoencoders, are difficult to train, and while some guidelines on hyperparameter optimization exist, one has to experiment with different settings. A systematic grid-search has not been carried out due to the vast number of parameters and exponentially scaling combinations of those. In fact, recently services have been found that support researchers and developers when performing such experiments (e.g. `comet.ml`). There are a number of architectural hyperparameters that can be adjusted: The architecture of the encoder (number of layers and their respective size), regularization, activation functions, optimization, and loss functions. In addition, during training, we have to decide on batch size, shuffling, and the number of epochs.

Our input data has 11 variables and we decide not to have more nodes than this number in any layer. Thereby we don't have to address challenges of overcomplete layers that can lead to the model simply passing the inputs without any learning happening (Vincent et al., 2010). Figure 14 summarizes the relatively simple network architecture that achieved good performance: The first two dense layers comprise the encoder, while the last two decode the data. In the encoded latent layer, the data is compressed down to 4 dimensions. We use a combination of hyperbolic tangent (tanh, `dense_5, dense_7`) and rectified linear unit (ReLU, `dense_6, dense_8`) activation functions. Again, this combination is not fully theory based but has shown good performance with autoencoder models more generally. For the sake of simplicity, we started with a mean-squared-error as our loss function/reconstruction error. However, experiments with various loss functions (Kullback-Leibler divergence, Poisson, Cosine Proximity), found Cosine Proximity loss and the Adam optimizer to deliver the best results. To prevent overfitting – in our case most likely leading to high reconstruction errors for any inputs previously unseen by the model – we introduce activity regularization in the first dense layer. We use L2 regularization, which is a method to penalize weights with large magnitudes, that are an indicator for overfitting (Ng, 2004). The model gets "smoothed out", which should make it more alert to anomalies.

The data is devided into 80% for training and 20% for validation. All anomalies are removed from the training set, leaving 1,715,843 observations. The network is trained in batches of 512 observations which are reshuffled in each epoch. We found that training the relatively small network for 10 epochs has been enough (we experimented with values up to 100 epochs). The training converged fast with no major accuracy gains being achieved after 2 epochs. Training time for one epoch on the Google Colaboratory GPU Engine averaged at 27 seconds. Given the very small model (only 223 trainable

---

[22]Variational autoencoers are a slightly more modern and intersting take on this class of models which also performed well in our experiments. Following the KISS principle, we decided to use the more traditional and simpler autoencoder architecture that is easier to explain and performed almost equally well.

parameters – weights and biases), the performance bottleneck was actually not the neural net computation itself and computation times dropped to 16 seconds on a 4 Core CPU machine, showing that in our case a GPU infrastructure was not necessary.

Once trained, the autoencoder can be used to reconstruct the test set of 431.659 observations, 2722 of which, approx 0.6% are our cases of interest. Here, all observations (normal and anomalies) are fed to the model and the reconstruction error is calculated as the Euclidean Distance, where we can clearly observe in Figure 17 the difference between the classes.

Figure 16: Density distribution of the reconstruction error, breakthrough patents in green

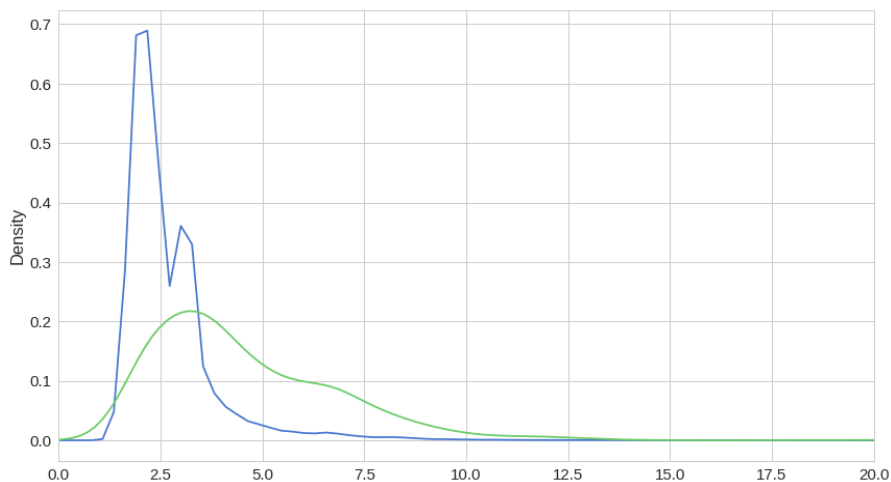

The error-term can be further used to calculate the ROC and AUC indicators. We achieve an intermediate AUC value of 0.8, which is not excellent but for this application reasonably good. We refer to this value as intermediate because the calculated error term functions as an ex-post parameter that can be arbitrarily set depending on the application. We can, for instance, decide to classify all estimations with an error between 3.8 and 11 to be considered breakthrough patents.

This would leave us with 1402 correctly identified breakthrough patents, a ROC-AUC value of 0.71. While far from an excellent result, this is a good result for this application. Not only is this a proof of concept but it also shows that breaktrough patents expose some significantly distinctive patterns.

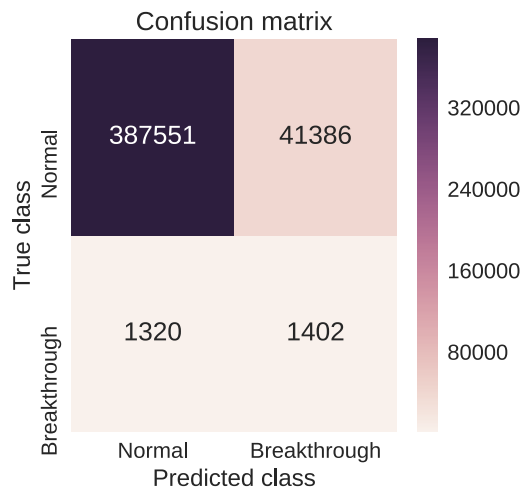

Figure 17: Confusion Matrix, error [3.8, 11]

# 4 Conclusion, ways forward, and avenues for future research

In this paper, we introduced the readership to the main idea behind predictive modeling, which somewhat stands in stark contrast to common intuition, workflows and data analysis routines of a trained econometricians developing causal models. Particularly, we elaborate on central concepts such as the establishment of generalizability via out-of-sample validation techniques. As a "bonus", we finally to a deep learning based workflow for anomaly detection, that can be used in cases when extremely rare observations need to be identified – the proverbial needle in the haystack problem.

Predictive modeling offers a promising methodology that can be used in diverse settings and is gaining relevance in a big data world. In this exercises our analysis stopped with the predicted results and their validation. We argue that scholars within the wider innovation studies and entrepreneurship community can adapt many approaches developed in ML, and in the following point towards some areas where we see the greatest potentials. These are (i.)the generation of quality indicators to quantify complex and up to now often unmeasurable theoretical concepts, (ii.) understanding the nature of rare events, (iii.) exploratory phenomenon spotting, and (iv.) the improvement of traditional statistical models particularly with explicit out-of sample evaluation.

To start with, we see great potential in employing predictions from a ML architecture as independent variables for up to now unobservable qualitative measures in a traditional regression setting. Such combinations of prediction and causal inference techniques offer the potential for granular and timely analysis of phenomena which currently cannot, or only to a limited extent, be addressed using traditional techniques and data sources. The "Startup Cartography Project" at the MIT (Andrews et al., 2017; Fazio et al., 2016; Guzman and Stern, 2015, 2017) provides a good example of such efforts. Coining it "nowcasting" and "placecasting", the project uses large amounts of business registration records and predictive analytics to estimate entrepreneurial quality for a substantial portion of registered firms in the US (about 80%) over 27 years. When engaging in such "predictions in the service of estimation" (Mullainathan and Spiess, 2017), it is not hard to see how these predictions of start-up quality (and similar quality indicators) might serve as dependent or independent variables in many interesting hypothesis-testing settings.

Related, one task our traditional econometrics toolkit has performed particularly bad, is the explanation but also prediction of extremely *rare events*, as we demonstrated in our empirical exercise. However, being able to explain impactful low probability events (also coined as "black swans, cf." Taleb, 2010) such as which start-up is going to be the next gazelle, which technology our patent is going to be the futures "next big thing", when does the next financial crisis hit or firm defaults (cf. e.g. van der Vegt et al., 2015), and so forth, would certainly be of enormous interest for research, policy, and business alike.

Along that line, a predictive model can be deployed in a more exploratory way for "phenomena spotting", to test if "there is something in the data". In our example, the anomaly detection part indicated that for over half of the breakthrough patents, there seem to exist some latent divergent patterns that the model picks up and that may be worth exploring for potential causality.

Lastly, the practice of out-of-sample testing might help to improve the external validity of our models by explicitly testing how good our model performs in terms of parameter estimates and overall prediction (Athey and Imbens, 2017).

We hope that this paper is also a contribution to initiating and shaping a dialogue between the disciplines. While the ML community may be ahead within predictive modelling of big data, applied econometricians have developed a deep understanding of causality, which is currently mostly lacking among ML practitioners. Social sciences have also a much richer tradition thinking about epistemology and – ideally – sampling, research design, ethics and arguably context-awareness. Such an interaction may be rather helpful when discussing current issues of algorithmic bias and explainable AI.

# A Appendix

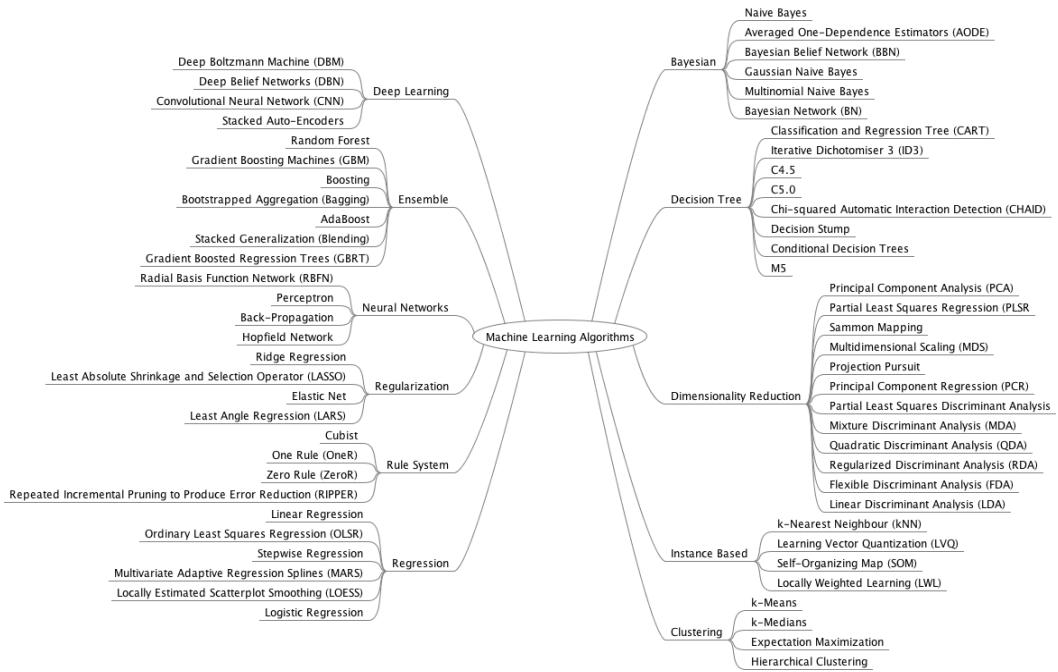

Figure A.1: Map of machine learning classes, techniques, and algorithms